\documentclass[11pt,a4paper]{article}
\usepackage[hyperref]{acl2019}
\usepackage{times}
\usepackage{latexsym}
\usepackage{afterpage}
\usepackage[utf8]{inputenc}
\usepackage{subcaption}

\usepackage{url}

\usepackage{graphicx}
\usepackage{mathtools}

\aclfinalcopy % Uncomment this line for the final submission
\def\aclpaperid{118} %  Enter the acl Paper ID here

\DeclareMathOperator{\mean}{\mu}
\DeclareMathOperator{\std}{\sigma}

\title{STRASS: A Light and Effective Method for Extractive Summarization \\Based on Sentence Embeddings}
  
\author{Léo Bouscarrat, 
  Antoine Bonnefoy, 
  Thomas Peel, 
  Cécile Pereira \\
  EURA NOVA \\
  Marseille, France \\
  \texttt{\{leo.bouscarrat,antoine.bonnefoy,}\\
  \texttt{thomas.peel,cecile.pereira\}@euranova.eu}}

\date{24/04/2019}

\begin{document}
\maketitle
\begin{abstract}
This paper introduces STRASS: Summarization by TRAnsformation Selection and Scoring. It is an extractive text summarization method which leverages the semantic information in existing sentence embedding spaces. 
Our method creates an extractive summary by selecting the sentences with the closest embeddings to the document embedding. The model learns a transformation of the document embedding to minimize the similarity between the extractive summary and the ground truth summary.
%In this approach, we learn a transformation of the document embedding in order to maximize the similarity between the ground truth summary and the generated summary. The sentences of the 
%obtain a position close to the positions of the sentences forming a good extractive summary.
%In this method, embeddings are used to compare proximities of sentences to documents. The closest  and the most 'interesting' sentences are selected. 
As the transformation is only composed of a dense layer, the training can be done on CPU, therefore, inexpensive. Moreover, inference time is short and linear according to the number of sentences.
As a second contribution, we introduce the French CASS dataset, composed of judgments from the French Court of cassation and their corresponding summaries.
On this dataset, our results show that our method performs similarly to the state of the art extractive methods with effective training and inferring time. 
\end{abstract}

\begin{figure*}[ht!]
\centering
\includegraphics[width=0.9\textwidth]{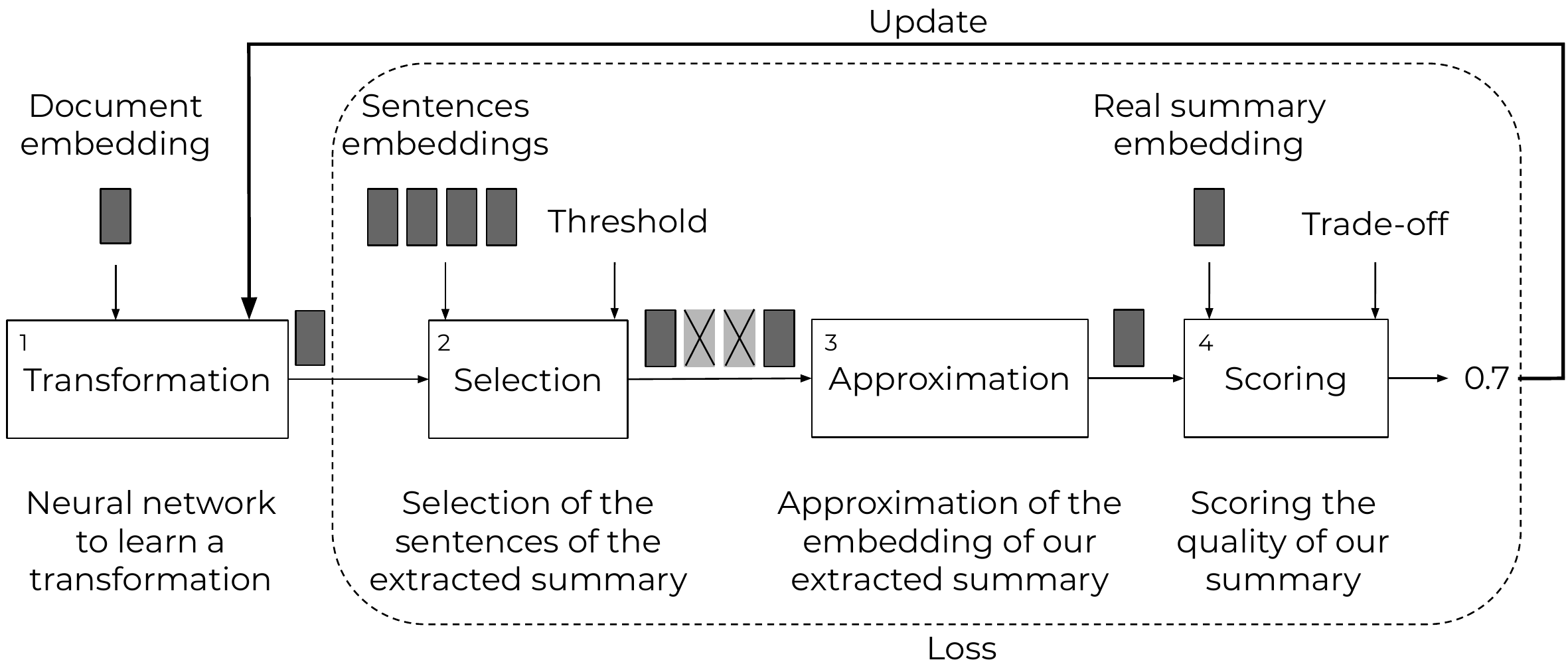}
\caption{Training of the model. The blocks present steps of the analysis.  All the elements above the blocks are inputs (document embedding, sentences embeddings, threshold, real summary embedding, trade-off).}
\label{fig:schema_model}
\end{figure*}

\section{Introduction}

Summarization remains a field of interest as numerous industries are faced with a growing amount of textual data that they need to process. Creating summary by hand is a costly and time-demanding task, thus automatic methods to generate them are necessary. There are two ways of summarizing a document: abstractive and extractive summarization.

In abstractive summarization, the goal is to create new textual elements to summarize the text. Summarization can be modeled as a sequence-to-sequence problem. For instance, \citet{D15-1044} tried to generate a headline from an article. However, when the system generates longer summaries, redundancy can be a problem. 
\citet{P17-1099} introduce a pointer-generator model (PGN) that generates summaries by copying words from the text or generating new words. Moreover, they added a coverage loss as they noticed that other models made repetitions on long summaries. 
Even if it provides state of the art results, the PGN is slow to learn and generate.
\citet{DBLP:journals/corr/PaulusXS17} added a layer of reinforcement learning on an encoder-decoder architecture but their results can present fluency issues.

In extractive summarization, the goal is to extract part of the text to create a summary. There are two standard ways to do that: a sequence labeling task, where the goal is to select the sentences labeled as being part of the summary, and a ranking task, where the most salient sentences are ranked first. 
It is hard to find datasets for these tasks as most summaries written by humans are abstractive.%Ways have been found to generate an extractive summary from an abstractive summary.
\citet{DBLP:journals/corr/NallapatiZZ16} introduce a way to train an extractive summarization model without labels by applying a Recurrent Neural Network (RNN) and using a greedy matching approach based on ROUGE.
Recently, \citet{N18-1158} combined reinforcement learning (to extract sentences) and an encoder-decoder architecture (to select the sentences).
%The use of the encoder-decoder architecture in this reinforcement model is different from the use in state of the art approaches as  abstractive models they generally use words as input whereas sentences.
%They use the ROUGE metrics for the rewards (section \ref{subsec:ROUGE}).

Some models combine extractive and abstractive summarization, using an extractor to select sentences and then an abstractor to rewrite them \cite{P18-1063, P18-1015, P18-1013}. They are generally faster than models using only abstractors as they filter the input while maintaining or even improving the quality of the summaries. 

% In this paper, we introduce a method of text summarization based on the use of pretrained sentence embeddings and their similarity.
% This summarization approach is cheap to use as it can be trained and generate summaries efficiently on CPU. 
% Je rajouterai la partie sur l'inférence quand j'aurai les résultats.
% The method is evaluated on a state of the art dataset in English and on a newly introduced dataset in French, the CASS dataset composed of judgments and their summaries written by lawyers.

%TODO lister les hypothèses de travail
%TODO lister les contributions

%Our goal is to generate extractive summaries. 
%We suppose that sentences that share close semantic meaning with an abstractive summary are good candidates for extractive summarization.   sentence and document embeddings retain semantic information that can be leverage in order to create extractive summaries. 

This paper presents two main contributions. 
First, we propose an inexpensive, scalable, CPU-trainable and efficient method of extractive text summarization based on the use of sentence embeddings. Our idea is that similar embeddings are semantically similar, and so by looking at the proximity of the embeddings it is possible to rank the sentences.
Secondly, we introduce the French CASS dataset (section \ref{subsec:datasets}), composed of 129,445 judgments with their corresponding summaries.

\section{Related Work}

%In extractive summarization, the task is to extract a subset of sentences from the text in order to create a summary. 
%The proposed summary composed of the extracted sentences should have a close semantic meaning with the reference summary. So we need a way to compute this semantic meaning.

In our model, STRASS, 
%it is unnecessary to learn an embedding function for the specific task of summarization and 
it is possible to use an embedding function \footnote{In this paper, `embedding function', `embedding space' and `embedding' will refer to the function that takes a textual element as input and outputs a vector, the vector space, and the vectors.}  trained with state of the art methods.

Word2vec is a classical method used to transform a word into a vector \cite{DBLP:journals/corr/abs-1301-3781}. 
Methods like word2vec keep information about semantics \cite{N13-1090}.
%So we need to look at embedding of sentences. 
%There are methods to create sentences and documents embedding functions such as doc2vec \cite{pmlr-v32-le14} and sent2vec \cite{DBLP:journals/corr/PagliardiniGJ17}. 
Sent2vec \cite{DBLP:journals/corr/PagliardiniGJ17} create embedding of sentences. It has state-of-the-art results on datasets for unsupervised sentence similarity evaluation.%, so it seems to be a useful tool to compare the similarity between our elements. 
%(the cosine similarity is used as the metric to compare the similarity between the embeddings of two sentences).

%In the remainder of this paper, the term 'embedding function' will be used to speak about the function that takes a textual element as input and outputs a vector, 'embedding space' for the space where the vectors are, and 'embedding' for the vectors.

EmbedRank \cite{K18-1022} applies sent2vec to extract keyphrases from a document in an unsupervised fashion. It hypothesizes that keyphrases that have an embedding close to the embedding of the entire document should represent this document well.
%So their first step is to extract all the candidates and then they choose the N candidates the closest of the documents. In order 
%To avoid repetition, they use MMR \cite{Carbonell:1998:UMD:290941.291025} to make a better selection of the candidates. MMR is a method that helps to reduce the redundancy.

We adapt this idea to select sentences for summaries (section \ref{subsec:baseline}). We suppose that sentences close to the document share some meaning with the document and are sentences that summarize well the text. We go further by proposing a supervised method where we learn a transformation of the document embedding to an embedding of the same dimension, but closer to sentences that summarize the text.% based on the hypothesis that even if summaries and documents have close semantic meanings, they have different embeddings.

\section{Model}

The aim is to construct an extractive summary. 
Our approach, STRASS, uses embeddings to select a subset of sentences from a document. 

We apply sent2vec to the document, to the sentences of the document, and to the summary. 
We suppose that, if we have a document with an embedding\footnote{Scalars are lowercased, vectors/embeddings are lowercased and in bold, sets are uppercased and matrices are uppercased and in bold.} $\mathbf{d}$ and a set $S$ with all the embeddings of the sentences of the document, and a reference summary with an embedding $\mathbf{ref\_sum}$, there is a subset of sentences $E_S \subset S$ forming the reference summary. Our target is to find an affine function $f(\cdot)$: ${\rm I\!R^\mathit{n}}\longrightarrow{\rm I\!R^\mathit{n}}$, such that:
\begin{equation*}
 \begin{cases}
    sim(\mathbf{s}, f(\mathbf{d})) \geq t  & \text{if } \mathbf{s} \in E_S \\
    sim(\mathbf{s}, f(\mathbf{d})) < t, & \text{otherwise}
  \end{cases}
\end{equation*}
Where $t$ is a threshold, and $sim$ is a similarity function between two embeddings. 

The training of the model is based on four main steps (shown in Figure \ref{fig:schema_model}):
\begin{itemize}
\item 
(1) Transform the document embedding by applying an affine function learned by a neural network (section \ref{subsec:neural_net});
\item
(2) Extract a subset of sentences to form a summary (section \ref{subsec:sentence_extrac});
\item
(3) Approximate the embedding of the extractive summary formed by the selected sentences (section \ref{subsec:embed_approx});
\item
(4) Score the embedding of the resulting summary approximation with respect to the embedding of the real summary (section \ref{subsec:final_scoring}).
\end{itemize}

To generate the summary, only the first two steps are used. The selected sentences are the output.
Approximation and scoring are only necessary during the training phase when computing loss function.

\subsection{Transformation}
\label{subsec:neural_net}

To learn an affine function in the embedding space, the model uses a simple neural network. A single fully-connected feed-forward layer. $f(\cdot)$: ${\rm I\!R^\mathit{n}}\longrightarrow{\rm I\!R^\mathit{n}}$:
\begin{equation*}
\begin{multlined}[t]
    f(\mathbf{d}) =  \mathbf{W} \times \mathbf{d} + \mathbf{b}
\end{multlined}
\end{equation*}
with \(\mathbf{W}\) the weight matrix of the hidden layer and \(\mathbf{b}\) the bias vector. Optimization is only conducted on these two elements.

\subsection{Sentence Extraction}
\label{subsec:sentence_extrac}

Inspired by EmbedRank~\cite{K18-1022} our proposed approach is based on embeddings similarities. Instead of selecting the top \(n\) elements, our approach uses a threshold. All the sentences with a score above this threshold are selected. As in \citet{DBLP:journals/corr/PagliardiniGJ17}, our similarity score is the cosine similarity. Selection of sentences is the first element:
\begin{equation*}
\begin{multlined}[t]
    sel(\mathbf{s}, \mathbf{d}, S, t) =  \\
sigmoid(ncos^{+}(\mathbf{s}, f(\mathbf{d}), S) - t)
\end{multlined}
\end{equation*}
with \(sigmoid\) the sigmoid function and \(ncos^{+}\) a normalized cosine similarity explained in section \ref{subsec:normalization}. A sigmoid function is used instead of a hard threshold as all the functions need to be differentiable to make the back-propagation. \(Sel\) outputs a number between 0 and 1. 1 indicates that a sentence should be selected and 0 that it should not. With this function, we select a subset of sentences from the text that forms our generated summary.

\subsection{Approximation}
\label{subsec:embed_approx}

As we want to compare the embedding of our generated extractive summary and the embedding of the reference summary, the model approximates the embedding of the proposed summary. 
%This part depends on the type of embedding used in the system. 
As the system uses sent2vec, the approximation is the average of the sentences weighted by the number of words in each sentence. We have to apply this approximation to the sentences extracted with \(sel\), which compose our generated summary. The approximation is:
\begin{equation*}
    app(\mathbf{d}, S, t) = \sum_{\mathbf{s} \in S} \mathbf{s} \times nb\_w(\mathbf{s}) \times sel(\mathbf{s}, \mathbf{d}, S, t)
\end{equation*}
where, \(nb\_w(\mathbf{s})\) is the number of words in the sentence corresponding to the embedding \(\mathbf{s}\).

\subsection{Scoring}
\label{subsec:final_scoring}

The quality of our generated summary is scored by comparing its embedding with the reference summary embedding. 
%First, we only tried to use the cosine similarity between the two, but it has a tendency to generate long summaries. 
Here, the compression ratio is added to the score in order to force the model to output shorter summaries. The compression ratio is the number of words in the summary divided by the number of words in the document.
\begin{equation*}
\begin{multlined}[t]
    loss = \lambda \times \frac{nb\_w(\mathbf{gen\_sum})}{nb\_w(\mathbf{d})} + \\
    (1 - \lambda) \times cos\_sim(\mathbf{gen\_sum}, \mathbf{ref\_sum})
\end{multlined}
\end{equation*}
with \(\lambda\) a trade-off between the similarity and the compression ratio, \(cos\_sim(\mathbf{x}, \mathbf{y})\), \(\mathbf{x}, \mathbf{y} \in {\rm I\!R}^n\) the cosine similarity and \(\mathbf{gen\_sum} = app(\mathbf{d}, S, t)\). The user should note that \(\lambda\) is also useful to change the trade-off between the proximity of the summaries and the length of the generated one. A higher \(\lambda\) results in a shorter summary.

\subsection{Normalization}
\label{subsec:normalization}

To use a single selection threshold on all our documents, a normalization is applied on the similarities to have the same distribution for the similarities on all the documents. 
% We use a normalization on our proximity metric.
First, we transform the cosine similarity from \(({\rm I\!R}^n, {\rm I\!R}^n) \longrightarrow[-1, 1]\) to \(({\rm I\!R}^n, {\rm I\!R}^n) \longrightarrow[0, 1]\):
\begin{equation*}
    cos^{+}(\mathbf{x}, \mathbf{y}) = \frac{cos\_sim(\mathbf{x}, \mathbf{y}) + 1}{2}
\end{equation*}

Then as in \citet{mori2002information} the function is reduced and centered in \(0.5\):
\begin{equation*}
\begin{multlined}[t]
   rcos^{+}(\mathbf{x}, \mathbf{y}, X) = \\
   0.5 + \frac{cos^{+}(\mathbf{x}, \mathbf{y}) - \underset{\mathbf{x_k} \in X}{\mean}(cos^{+}(\mathbf{x_k}, \mathbf{y}))}{\underset{\mathbf{x_k} \in X}{\std}(cos^{+}(\mathbf{x_k}, \mathbf{y}))} \\
\end{multlined}
\end{equation*}
where \(\mathbf{y}\) is an embedding, \(X\) is a set of embeddings, \(\mathbf{x}  \in  X\), \(\mean\) and \(\std\) are the mean and standard deviation. 

A threshold is applied to select the closest sentences on this normalized cosine similarity.
%it will be better if we were sure that all the texts will have at least one sentence selected whatever is the threshold.
In order to always select at least one sentence, we restricted our similarity measure % and be sure that we always select at least one sentence. We restrict it 
in $(-\infty, 1]$, where, for each document, the closest sentence has a similarity of 1:
%There is no lower or upper bound in this distribution, but it is better if our score has at least one of those. So \(rcos^{+}\) is divided by the maximum of the distribution, so the best sentence has always a score of 1.
\begin{equation*}
\begin{multlined}[t] 
   ncos^{+}(\mathbf{x}, \mathbf{y}, X) = \frac{rcos^{+}(\mathbf{x}, \mathbf{y}, X)}{\underset{\mathbf{x_k} \in X}{\max}(rcos^{+}(\mathbf{x_k}, \mathbf{y}, X))}
\end{multlined}
\end{equation*}

\section{Experiments}

\subsection{Datasets}
\label{subsec:datasets}
To evaluate our approach, two datasets were used with different intrinsic document and summary structures which are presented in this section. More detailed information is available in the appendices (table \ref{table:1}, figure \ref{fig: info_gen_fr} and figure \ref{fig: info_gen_en}).

We introduce a new dataset for text summarization, the CASS dataset\footnote{The dataset is available here: \url{https://github.com/euranova/CASS-dataset}}. This dataset is composed of 129,445 judgments given by the French Court of cassation between 1842 and 2016 and their summaries (one summary by original document).
Those summaries are written by lawyers and explain in a short way the main points of the judgments. 
%The judgments are structured, however, each part of the structure can be more or less long and the information highlighted by the summary is not always at the beginning or the end. 
As multiple lawyers have written summaries, there are different types of summary ranging from purely extractive to purely abstractive. 
This dataset is maintained up-to-date by the French Government and new data are regularly added. Our version of the dataset is composed of 129,445 judgements.

The CNN/DailyMail dataset  \cite{Hermann:2015:TMR:2969239.2969428, K16-1028} is composed of 312,084 couples containing a news article and its highlights. The highlights show the key points of an article. We use the split created by \citet{K16-1028} and refined by \citet{P17-1099}. 
%Contrary to the French Court of cassation dataset, in the CNN/DailyMail dataset, there are articles in different domains such as sport, economics or politics. Therefore, different structures exist.

%The last dataset \cite{10.1007/978-3-642-32695-0_6} is composed of legal case reports from the Federal Court of Australia. We consider a summary to be the concatenation of the catchphrases of the report. A catchphrase is a word or a group of word linked to the case report. This dataset is composed of 3,890 documents.

\subsection{Baseline}
\label{subsec:baseline}
An unsupervised version of our approach is to use the document embedding as an approximation for the position in the embedding space used to select the sentences of the summary. It is the application of EmbedRank \cite{K18-1022} on the extractive summarization task.
This approach is used as a baseline for our model

\subsection{Oracles}
We introduce two oracles. Even if these models do not output the best possible results for extractive summarization, they show good results.

The first model, called $Oracle$, is the same as the baseline, but instead of taking the document embedding, the model takes the embedding of the summary and then extracts the closest sentences.

The second model, called $Oracle sent$, extracts the closest sentence to each sentence of the summary. This is an adaptation of the idea that \citet{DBLP:journals/corr/NallapatiZZ16} and \citet{P18-1063} used to create their reference extractive summaries.

\subsection{Evaluation details}
\label{subsec:ROUGE}
ROUGE \cite{W04-1013} is a widely used set of metrics to evaluate summaries. The three main metrics in this set are ROUGE-1 and ROUGE-2, which compare the 1-grams and 2-grams of the generated and reference summaries, and ROUGE-L, which measures the longest sub-sequence between the two summaries. ROUGE is the standard measure for summarization, especially because more sophisticated ones like METEOR \cite{W14-3348} require resources not available for many languages.

Our results are compared with the unsupervised system TextRank \cite{mihalcea2004textrank, DBLP:journals/corr/BarriosLAW16} and with the supervised systems Pointer-Generator Network \cite{P17-1099} and $rnn-ext$ \cite{P18-1063}. The Pointer-Generator Network is an abstractive model and $rnn-ext$ is extractive.

For all datasets, a sent2vec embedding of dimension 700 was trained on the training split. To choose the hyperparameters, a grid search was computed on the validation set. Then the set of hyperparameters with the highest ROUGE-L were used on the test set. The selected hyperparameters are available in appendix \ref{subsec:hyperparam}.

\begin{table}[ht!]
\centering
\begin{tabular}{ | c | c | c | c | }
    \hline
    & R1 F1 & R2 F1 & RL F1 \\
    \hline
    Baseline & 39.57 & 22.11 & 29.71 \\
    TextRank & 39.30 & 23.49 & 31.45 \\
    \hline 
    PGN & \textbf{53.25} & \textbf{40.25} & \textbf{45.58} \\
    rnn-ext & \textbf{53.05} & 38.21 & 44.62 \\
    STRASS & 52.68 & 38.87 & 44.72 \\
    \hline
    Oracle & 62.79 & 50.10 & 55.03 \\
    Oracle sent & 63.90 & 50.56 & 55.75 \\
    \hline
\end{tabular}
\caption{Results of different models on the French CASS dataset using ROUGE with 95\% confidence. The models of the first block are unsupervised, the models of the second block are supervised and the models of the last block are the oracles. F1 is the F-measure. R1, R2 and RL stand for ROUGE1, ROUGE2, and ROUGE-L.}
\label{table:cass_small}
\end{table}

\begin{table}[ht!]
\centering
\begin{tabular}{ |c | c | c | c | }
    \hline
    & R1 F1 & R2 F1 & RL F1 \\
    \hline
    Baseline & 34.02 & 12.48 & 28.27 \\
    TextRank & 30.83 & 13.02 & 27.39 \\
    \hline
    PGN* & 39.53 & 17.28 & \textbf{36.38} \\
    rnn-ext* & \textbf{40.17} & \textbf{18.11} & \textbf{36.41} \\
    STRASS & 33.99 & 14.18 & 30.04 \\
    \hline
    Oracle & 43.55 & 22.43 & 38.47 \\
    Oracle sent & 46.21 & 25.81 & 42.47 \\
    Lead3 & 40.00 & 17.56 & 36.33 \\
    Lead3 - PGN* & 40.34 & 17.70 & 36.57 \\
    \hline
\end{tabular}
\caption{Results of different models on the CNN/DailyMail. The Lead3 - PGN is the lead 3 score as reported in \cite{P17-1099}. The scores with * are taken from the corresponding publications. F1 is the F-measure. R1, R2 and RL stand for ROUGE1, ROUGE2, and ROUGE-L.}
\label{table:cnn_dm_small}
\end{table}

%\begin{table}[ht!]
%\centering
%\begin{tabular}{ |c | c | c | c | }
%    \hline
%    & R1 F1 & R2 F1 & RL F1 \\
%    \hline
%    Baseline & 23.09 & 6.31 & 20.22 \\
%    TextRank & 21.06 & 6.11 & 18.81 \\
%    \hline
%    Model & 25.40 & 7.78 & 22.98 \\
%    \hline
%    Oracle & 35.78 & 15.39 & 32.56 \\
%    \hline
%\end{tabular}
%\caption{Results of different models on the Australian legal case reports dataset.}
%\label{table:au_small}
%\end{table}

\begin{figure}[ht!]
\centering
\includegraphics[width=0.45\textwidth]{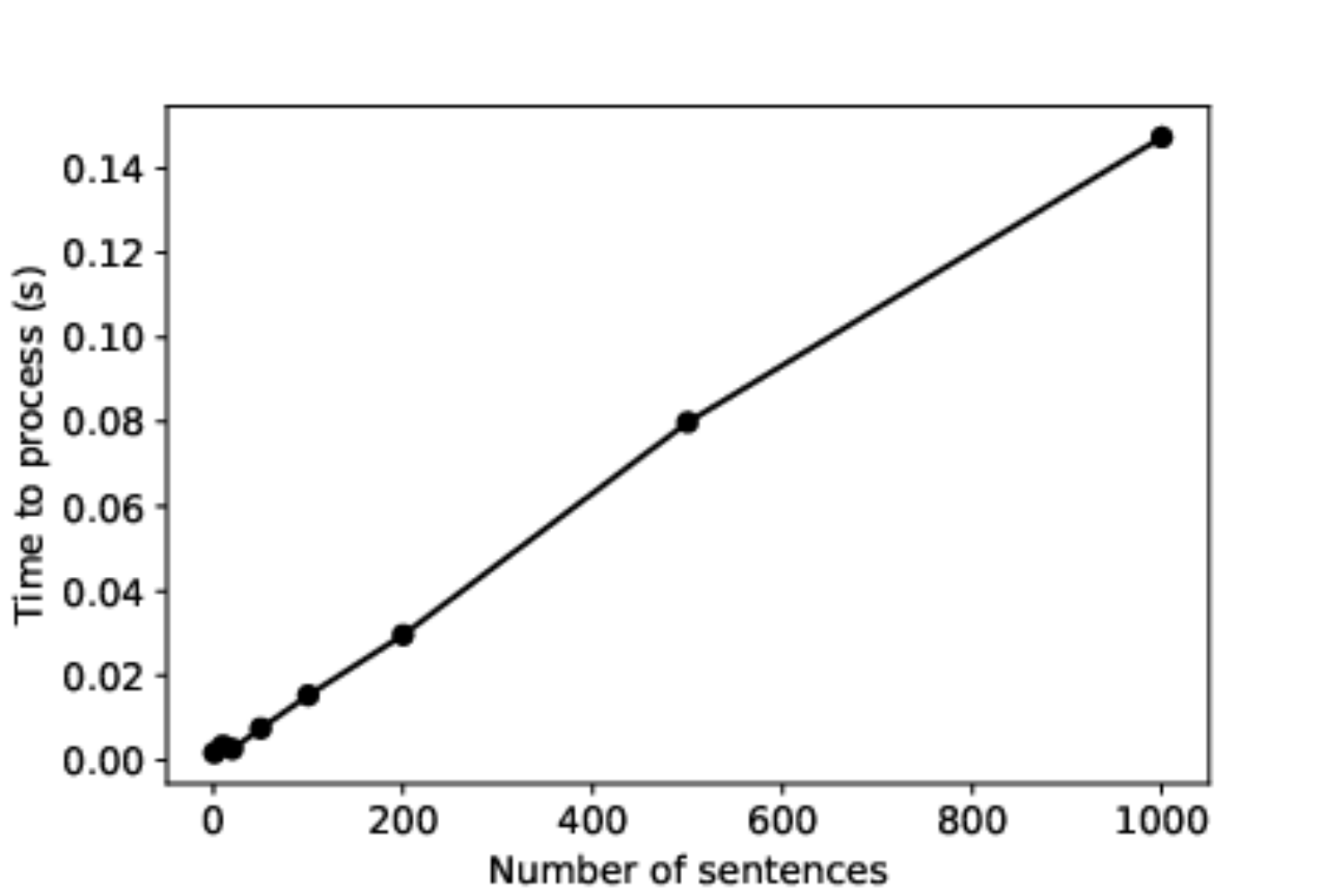}
\caption{Processing time of the summarization function (y-axis) by the number of lines of the text as input (x-axis). Results computed on an i7-8550U.}
\label{fig:time_by_lines}
\end{figure}

\section{Results}

Tables \ref{table:cass_small} and \ref{table:cnn_dm_small} present the results for the CASS and the CNN/DailyMail datasets.
As expected, the supervised model performs better than the unsupervised one. On the three datasets, the supervision has improved the score in terms of ROUGE-2 and ROUGE-L. In the same way, our oracles are always better than the learned models, proving that there is still room for improvements. Information concerning the length of the generated summaries and the position of the sentences taken are available in the appendices \ref{sec:words_sent}.

On the French CASS dataset, our method performs similarly to the $rnn-ext$. The PGN performs a bit better (+0.13 ROUGE-1, +0.38 ROUGE-2, + 0.81 ROUGE-L compared to the other models), which could be linked to the fact that it can select elements smaller than sentences.

On the CNN/DailyMail dataset, our supervised model performs poorly. 
We observe a significant difference (+2.66 ROUGE-1, +3.38 ROUGE-2, and +4.00 ROUGE-L) between the two oracles. It could be explained by the fact that the summaries are multi-topic and our models do not handle such case.
Therefore, as our loss doesn't look at the diversity, STRASS may miss some topics in the generated summary.

%discussed in the summary, so the interesting sentences might be far in the embedding space. 
A second limitation of our approach is that our model doesn't consider the position of the sentences in the summary, information which presents a high relevance in the CNN-Dailymail dataset. 
%The models with RNNs have information about temporality. To mimic that, models such as \citet{narayan-etal-2018-dont} adds position embeddings to add such information.
%On the Australian legal case reports dataset, even with fewer data available (only 3,096 documents in the train set), our supervised model show increased results (+2.31 ROUGE-1, +1.47 ROUGE-2, +2.76 ROUGE-L) when compared to the baseline.

STRASS has some advantages. 
First, it is trainable on CPU and thus light to train and run. 
Indeed, the neural network in our model is only composed of one dense layer. The most recent advances in text summarization with neural networks are all based on deep neural networks requiring GPU to be learned efficiently.
Second, the method is scalable. The processing time is linear with the number of lines of the documents (Figure \ref{fig:time_by_lines}). The model is fast at inference time as sent2vec embeddings are fast to generate. Our model generated the 13,095 summaries of the CASS dataset in less than 3 minutes on an i7-8550U CPU. 

%Third, as the model uses only embeddings, there is no issue of language as it is possible to either train an embedding model on the dataset or to use a pretrained model.

\section{Conclusion and Perspectives}

%Our unsupervised pipeline shows also an interesting result when competing with TextRank, a more in-depth study of this method compared to state of the art unsupervised could be interesting.

%Our oracle shows that improvements are still possible and we are far of the best result possible.
%We observed that, in most of the cases, our oracles, even if resulting in 100\% extractive summaries, allow to obtain summaries with better rouge scores than the state of the art methods. 
To conclude, we proposed here a simple, cost-effective and scalable extractive summarization method. 
STRASS creates an extractive summary by selecting the sentences with the closest embeddings to the projected document embedding. The model learns a transformation of the document embedding to maximize the similarity between the extractive summary and the ground truth summary. 
We showed that our approach obtains similar results than other extractive methods in an effective way.

There are several perspectives to our work. 
%First, we want to learn the threshold instead of using the same on all the documents as it could allow better flexibility in terms of number of sentences extracted.
%We also want to use other information to learn the embedding of the summary, mainly 
First, we would like to use the sentence embeddings as an input of our model, as this should increase the accuracy.
Additionally, we want to investigate the effect of using other sent2vec embedding spaces (especially more generalist ones) or other embedding functions like doc2vec \cite{pmlr-v32-le14} or BERT \cite{devlin-etal-2019-bert}.

For now, we have only worked on sentences but this model can use any embeddings, so we could try to build summaries with smaller textual elements than sentences such as key-phrases, noun phrases...
Likewise, to apply our model on multi-topic texts, we could try to create clusters of sentences, where each cluster is a topic, and then extract one sentence by cluster. 

Moreover, currently, the loss of the system is only composed of the proximity and the compression ratio. Other meaningful metrics for document summarization such as diversity and representativity could be added into the loss. 
%By adding these elements to the loss, improvements could be made. 
Especially, submodular functions could (1) allow to obtain near-optimal results and (2) allow to include elements like diversity  \cite{Lin:2011:CSF:2002472.2002537}. Another information we could add is the position of the sentences in the documents like \citet{narayan-etal-2018-dont}.

Finally, the approach could be extended to query-based summarization \cite{2013MuraliKrishnaQueryBased}. One could use the embedding function on the query and take the sentences that are the closest to the embedding of the query.

%Positive points of our methods:
%* can be apply on any language:
%    * can retrain the embedding on the train dataset or use a pretrained embedding
%* Easily applicable on multi-document summarization
%* Easily applicable on request summarization (summarization around a request made by an user, the request just %need to be transform into an embedding)
%* Next work on the learning of the threshold ?
%* Next work on using also the embeddings of the sentences to approximate the embedding of the summary
%* Next work on unit smaller than sentences ?

\section*{Acknowledgement}

We thank Cécile Capponi, Carlos Ramisch, Guillaume Stempfel, Jakey Blue and our anonymous reviewer for their helpful comments. 

\bibliography{acl2019}
\bibliographystyle{acl_natbib}

%\afterpage{\null\newpage}
%\newpage

\appendix

\section{Appendices}
\label{sec:appendix}
%Appendices are material that can be read, and include lemmas, formulas, proofs, and tables that are not critical to the reading and understanding of the paper. 
%Appendices should be \textbf{uploaded as supplementary material} when submitting the paper for review. Upon acceptance, the appendices come after the references, as shown here. Use
%\verb|\appendix| before any appendix section to switch the section
%numbering over to letters.

\subsection{Datasets}

The composition of the datasets and the splits are available in table \ref{table:1}.

\subsection{Preprocessing}

On the French CASS dataset, we have deleted all the accents of the texts and we have lower-cased all the texts as some of them where entirely upper-cased without any accent. To create the summaries, all the ANA parts of the XML files provided in the original dataset where taken and concatenate to form a single summary for each document. These summaries explain different key points of the judgment.
On the CNN/DailyMail, the preprocessing of \citet{P17-1099} was used. As an extra cleaning step, we deleted the documents that had an empty story.

\subsection{Hyperparameters}
\label{subsec:hyperparam}

To obtain the embeddings functions for both datasets we trained a sent2vec model of dimension 700 with unigrams on the train splits.

For the CASS dataset, the baseline model has a threshold at $0.8$, the oracle at $0.8$ and STRASS has a threshold at $0.8$ and a $\lambda$ at $0.3$. TextRank was used with a ratio of $0.2$. The PGN
For the CNN/DailyMail dataset, the baseline model has a threshold at $1.0$, the oracle at $0.9$ and STRASS has a threshold at $0.8$ and a $\lambda$ at $0.4$. TextRank was used with a ratio of $0.15$.

%For the Australian legal case reports dataset, the baseline model has a threshold at $0.95$, the oracle at $0.9$ and the model has a threshold at $0.8$ and a $\lambda$ at $0.4$.

\subsection{Results}

\subsubsection{ROUGE Score}

More detailed results are available in tables \ref{table:4} and \ref{table:5}. High recall with low precision is generally synonym of long summary.

\begin{table*}[ht!]
    \centering
    \begin{tabular}{| c | c | c | c | c | c | c | c |}
    \hline
         Dataset & $s_d$ & $s_s$ & $t_d$ & $t_s$ & train & val & test \\
         \hline
         CASS & 19.4 & 1.6 & 894 & 114 & 103,434 & 12,916 & 13,095 \\
         CNN/DailyMail & 28.9 & 3.8 & 786 & 53 & 287,112 & 13,367 & 11,489\\
         %Legal Case AU & 294.1 & 8.0 & 6,390 & 64 & 3,096 & 388 & 402 \\
    \hline
    \end{tabular}
    \caption{Size information for the datasets, $s_d$ and $s_s$ are respectively the average number of sentences in the document and in the summary, $t_d$ and $t_t$ are respectively the number of tokens in the document and in the summary. train, val and test are respectively the number of documents in the train, validation and test sets.}
    \label{table:1}
\end{table*}

\begin{table*}[ht!]
\centering
\begin{tabular}{ | c | c | c | c | c | c | c | c | c | c | }
    \hline
    & R1 P & R1 R & R1 F1 & R2 P & R2 R & R2 F1 & RL P & RL R & RL F1 \\
    \hline
    Baseline & 32.27 & 65.81 & 39.55 & 17.98 & 36.88 & 22.09 & 24.13 & 50.11 & 29.69 \\
    TextRank & 32.62 & 68.58 & 39.30 & 19.47 & 41.95 & 23.49 & 25.98 & 56.16 & 31.45 \\
    \hline 
    PGN & 69.70 & 49.01 & 53.25 & 53.46 & 36.67 & 40.25 & 60.31 & 41.65 & 45.58 \\
    rnn-ext & 49.54 & 69.62 & 53.12 & 35.94 & 50.00 & 38.30 & 42.03 & 58.43 & 44.77 \\
    STRASS & 56.23 & 62.55 & 52.68 & 41.97 & 45.71 & 38.87 & 48.05 & 52.93 & 44.72
    \\
    \hline
    Oracle & 66.40 & 68.41 & 62.79 & 53.80 & 53.40 & 50.10 & 58.73 & 59.20 & 55.03 \\
    Oracle sent & 69.50 & 64.82 & 63.90 & 55.36 & 50.91 & 50.56 & 60.77 & 56.34 & 55.75 \\
    \hline
\end{tabular}
\caption{Full results of different models on the French CASS dataset using ROUGE with 95\% confidence. The models in the first part are unsupervised models, then supervised models and the last part is the oracle. P is precision, R is recall and F1 is the F-measure. R1, R2 and RL stand for ROUGE1, ROUGE2, and ROUGE-L.}
\label{table:4}
\end{table*}

\begin{table*}[ht!]
\centering
\begin{tabular}{ | c | c | c | c | c | c | c | c | c | c | }
    \hline
    & R1 P & R1 R & R1 F1 & R2 P & R2 R & R2 F1 & RL P & RL R & RL F1 \\
    \hline
    Baseline & 32.67 & 40.09 & 34.02 & 12.07 & 14.65 & 12.48 & 27.29 & 33.14 & 28.27 \\
    TextRank & 23.44 & 59.29 & 30.83 & 9.95 & 25.02 & 13.02 & 20.77 & 53.02 & 27.39 \\
    \hline
    PGN* &  &  & 39.53 &  &  & 17.28 &  &  & 36.38 \\
    rnn-ext* &  &  & 40.17 &  &  & 18.11 &  &  & 36.41 \\
    STRASS & 28.56 & 53.53 & 33.99 & 11.89 & 22.62 & 14.18 & 25.21 & 47.46 & 30.04 \\
    \hline
    Oracle & 44.92 & 50.93 & 43.55 & 24.14 & 25.47 & 22.43 & 39.98 & 44.75 & 38.47 \\
    Oracle sent & 35.17 & 74.30 & 46.21 & 19.84 & 40.83 & 25.81 & 32.37 & 68.12 & 42.47 \\
    Lead3 & 33.89 & 53.35 & 40.00 & 14.84 & 23.59 & 17.56 & 30.80 & 48.43 & 36.33 \\
    Lead - PGN* &  &  & 40.34 &  &  & 17.70 &  &  & 36.57 \\
    \hline
\end{tabular}
\caption{Full results of different models on the CNN/DailyMail. The Lead3 - PGN is the lead 3 score as reported in \cite{P17-1099}. The scores with * are taken from the corresponding publications. P is precision, R is recall and F1 is the F-measure. R1, R2 and RL stand for ROUGE1, ROUGE2, and ROUGE-L.}
\label{table:5}
\end{table*}

\afterpage{\null\newpage}

\begin{figure*}[ht!]
    \centering
    
    \begin{subtable}[b]{0.45\textwidth}
        \centering
        \begin{tabular}{| c | c | c | c |}
            \hline
             Model & $s$ & $w$ & $w/s$ \\
             \hline
             Reference & 1.6 & 117 & 73.1 \\
             STRASS & 2.0 & 151 & 75.5 \\
             Oracle & 1.7 & 138 & 81.2 \\
             Oracle sent & 1.5 & 112 & 74.7 \\
            \hline
        \end{tabular}
        \caption{Size information for the generated summary on the test split of the CASS dataset, $s$, $w$, $w/s$ are respectively the average number of sentences, the average number of words and the average number of words per sentences.}
        \label{table:fr_result_len}
    \end{subtable}
    ~
    \begin{subfigure}[b]{0.45\textwidth}
        \includegraphics[width=\textwidth]{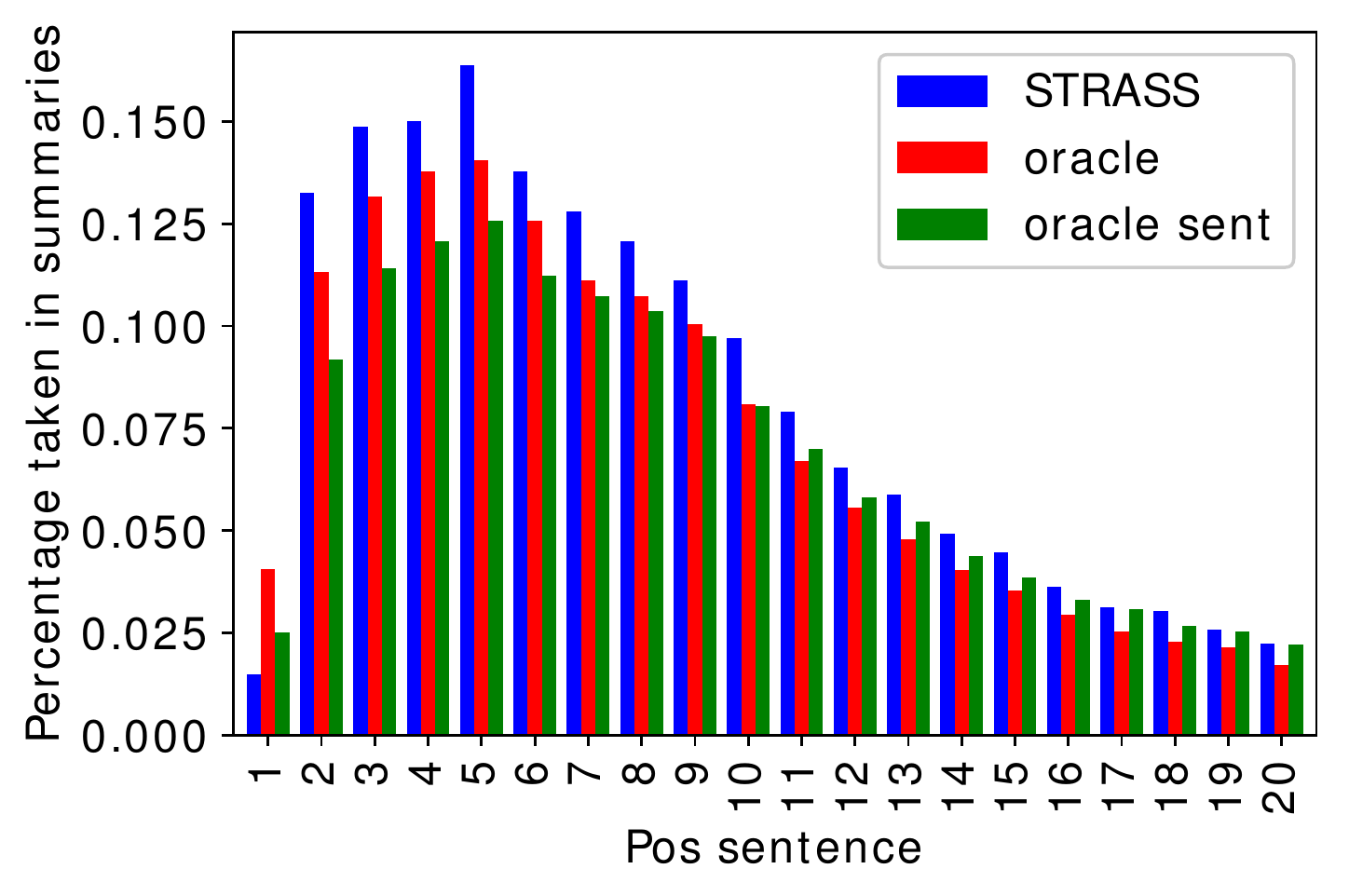}
        \caption{Percentage of times that a sentence is taken in a generated summary in function of their position in the document on the CASS dataset.}
        \label{fig:pos_sum20_fr_model_test}
    \end{subfigure}
    
    \centering
    
    \begin{subfigure}[b]{0.45\textwidth}
        \includegraphics[width=\textwidth]{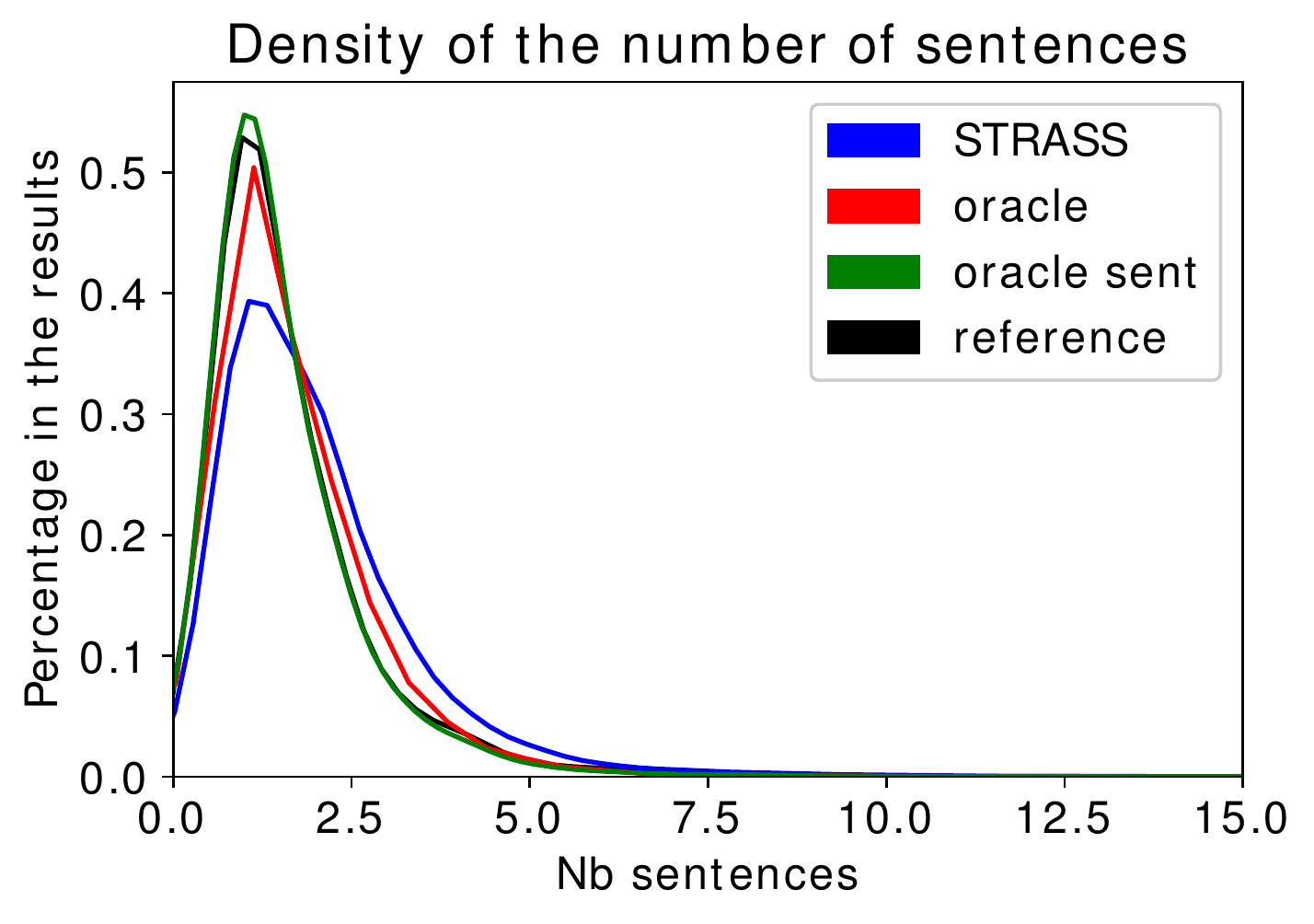}
        \caption{Density of the number of sentences in the generated summaries for several models and the reference on the CASS dataset.}
        \label{fig:density_sent_fr_test}
    \end{subfigure}
    ~
    \begin{subfigure}[b]{0.45\textwidth}
        \includegraphics[width=\textwidth]{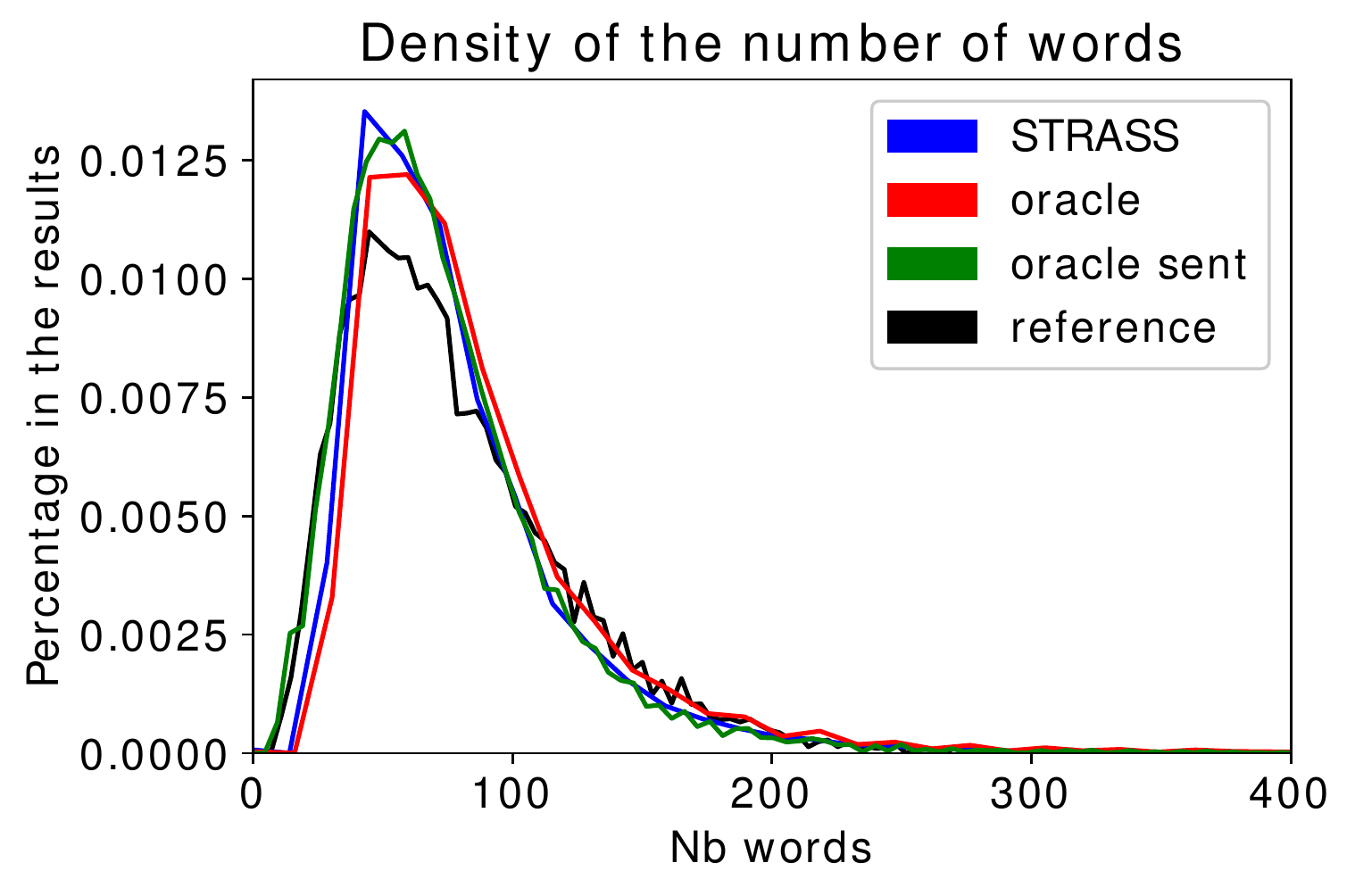}
        \caption{Density of the number of words in the generated summaries for several models and the reference on the CASS dataset.}
        \label{fig:density_word_fr}
    \end{subfigure}
    \caption{Information about the length of the generated summaries for the CASS dataset.}
    \label{fig: info_gen_fr}
\end{figure*}

\afterpage{\null\newpage}
\newpage

\subsubsection{Words and sentences}
\label{sec:words_sent}

On the French CASS dataset the summaries generated by the models are generally close in terms of length (number of words, number of sentences and number of words per sentences (figure \ref{table:fr_result_len}, \ref{fig:density_sent_fr_test}, \ref{fig:density_word_fr})). All the tested extractive methods tend to select sentences at the beginning of the documents. The first sentence make an exception to that rule (figure \ref{fig:pos_sum20_fr_model_test}). We observe that this sentence can have the list of the lawyers and judges that were present at the case. STRASS tends to generate longer summaries with more sentences. The discrepancy in the average number of sentences between the reference and \(Oracle sent\) is due to sentences that are extracted multiple times.

\begin{figure*}[ht!]
    \centering
    
    \begin{subtable}[b]{0.45\textwidth}
        \centering
        \begin{tabular}{| c | c | c | c |}
        \hline
             Model & $s$ & $w$ & $w/s$\\
             \hline
             Reference & 3.9 & 55 & 14.1\\
             STRASS & 2.7 & 135 & 50 \\
             Oracle & 1.5 & 84 & 56 \\
             Oracle sent & 3.5 & 137 & 39.1\\
        \hline
        \end{tabular}
        \caption{Size information for the generated summary on the test split of the CNN/DM dataset, $s$, $w$, $w/s$ are respectively the average number of sentences, the average number of words and the average number of words per sentences.}
        \label{table:en_result_len}
    \end{subtable}
    ~
    \begin{subfigure}[b]{0.45\textwidth}
        \includegraphics[width=\textwidth]{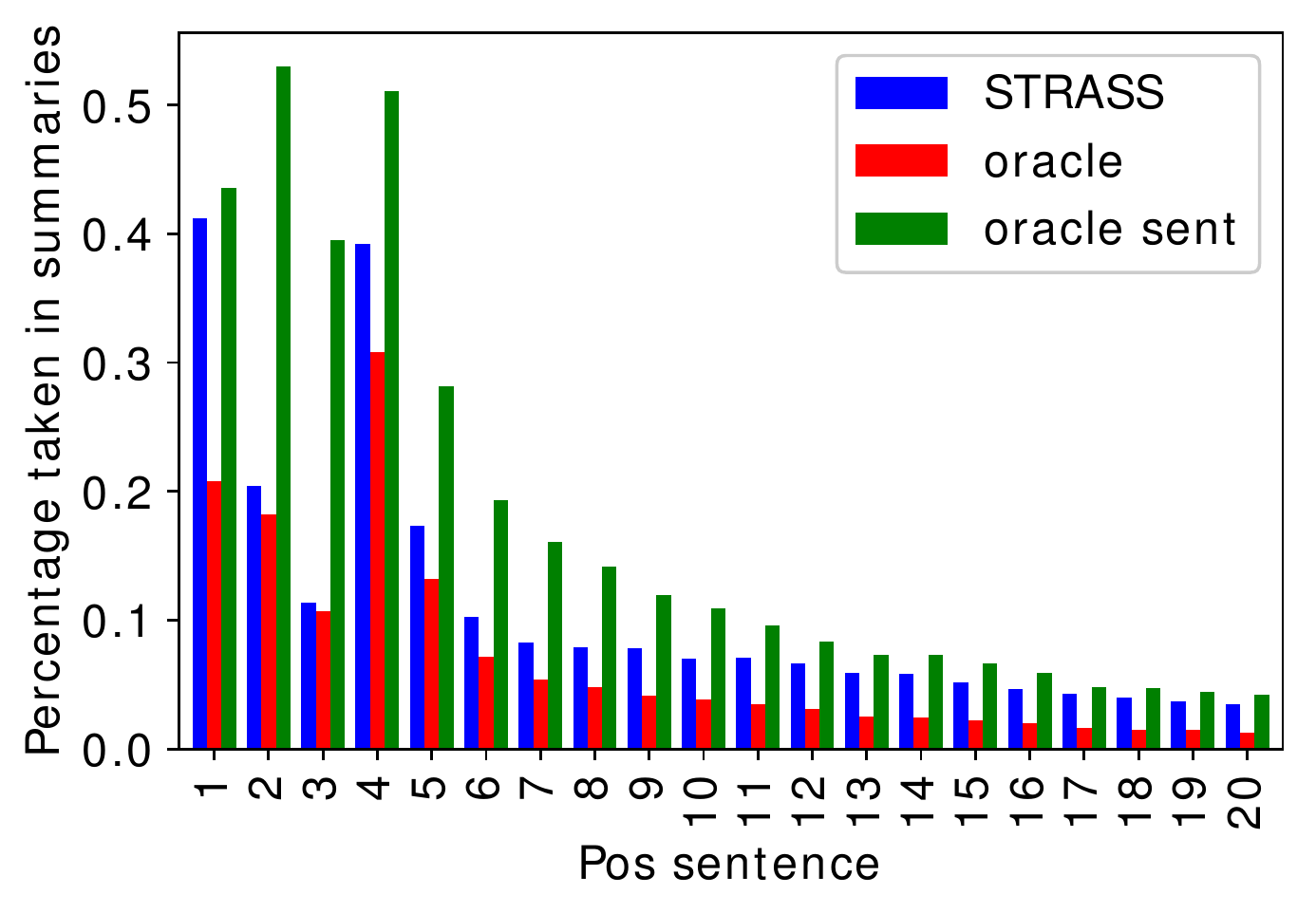}
        \caption{Percentage of times that a sentence is taken in a generated summary in function of their position in the document on the CNN/DM dataset.}
        \label{fig:pos_sum20_en_model}
    \end{subfigure}
    
    \centering
    
    \begin{subfigure}[b]{0.45\textwidth}
        \includegraphics[width=\textwidth]{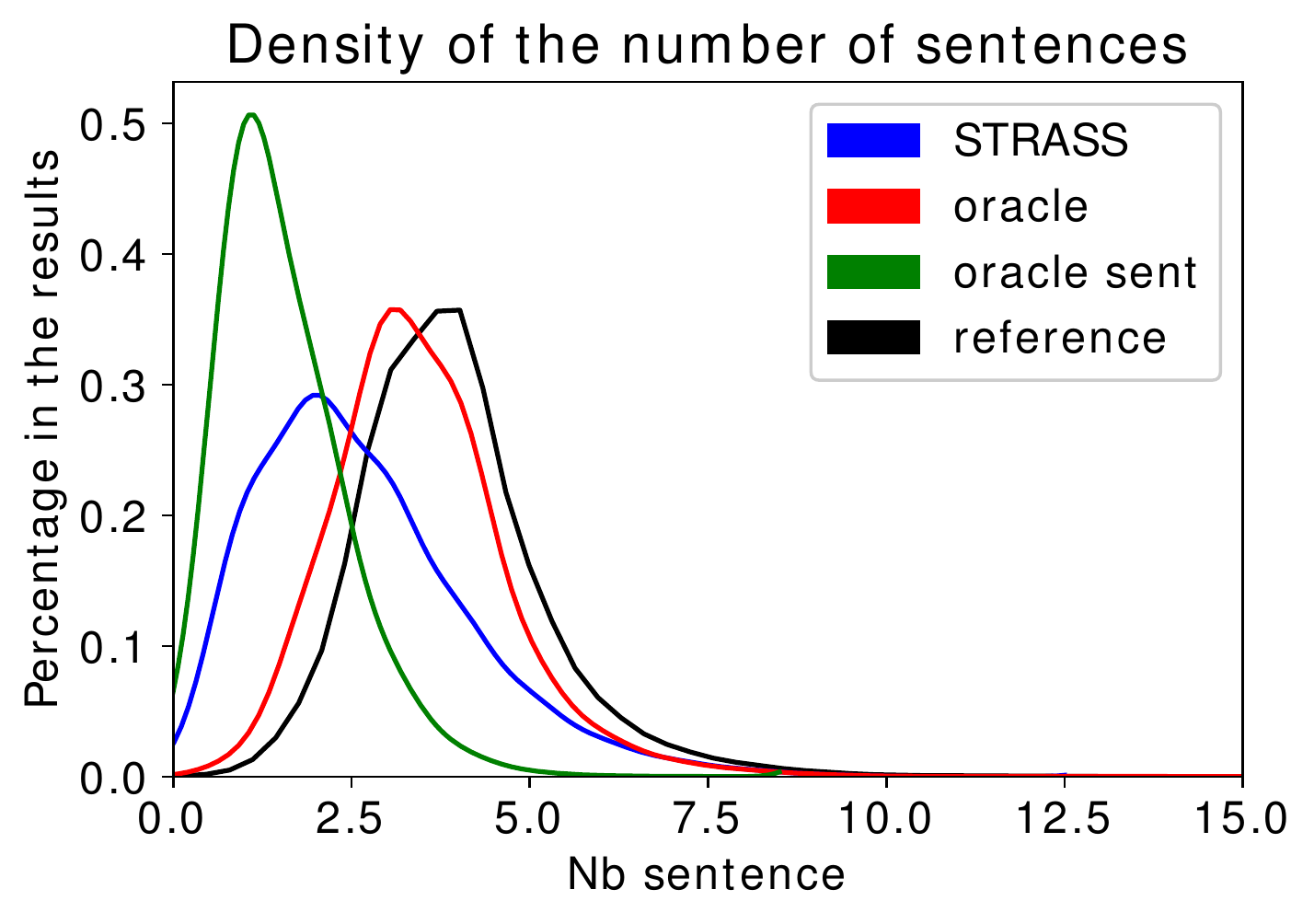}
        \caption{Density of the number of sentences in the generated summaries for several models and the reference on the CNN/DM dataset.}
        \label{fig:density_sent_en}
    \end{subfigure}
    ~
    \begin{subfigure}[b]{0.45\textwidth}
        \includegraphics[width=\textwidth]{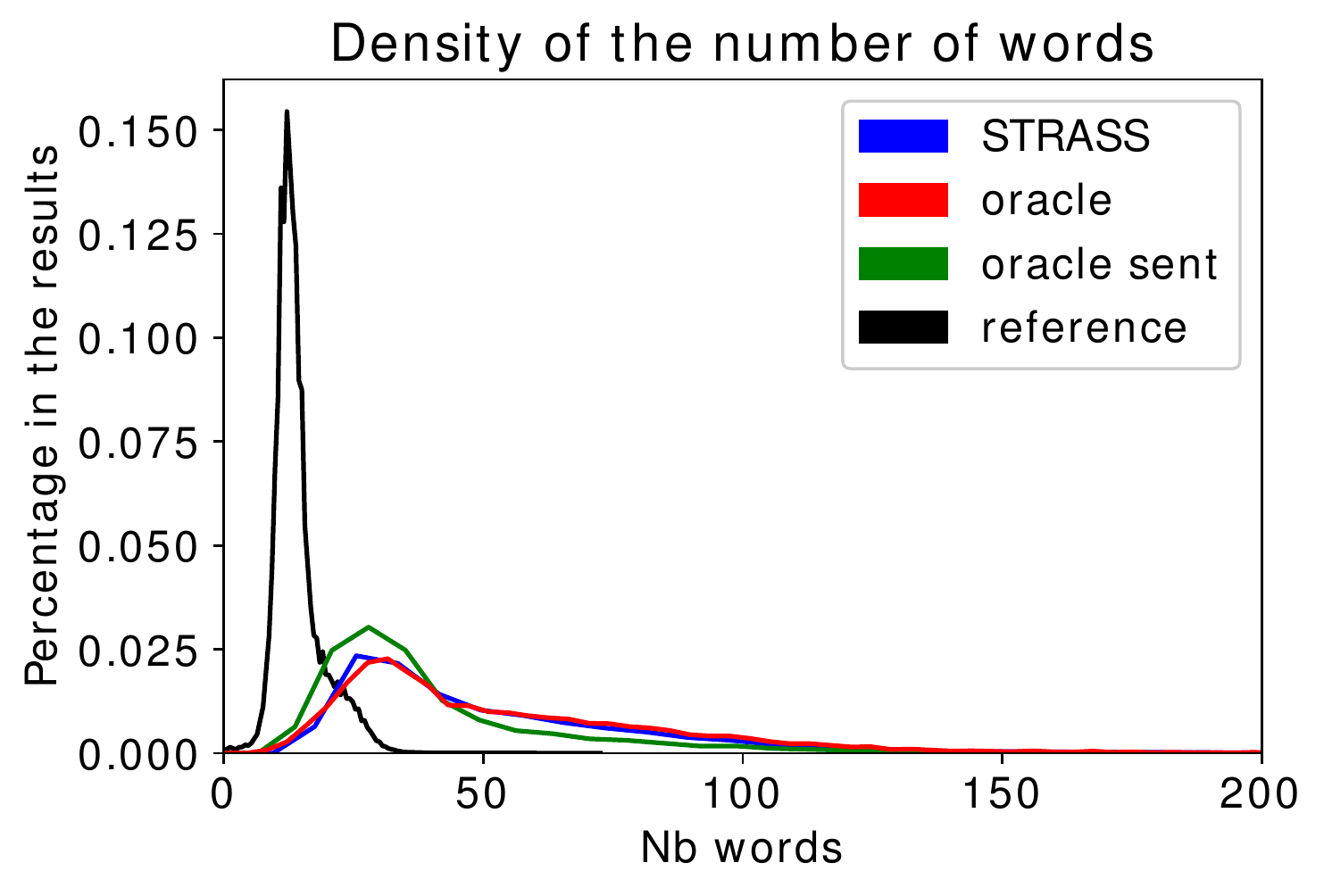}
        \caption{Density of the number of words in the generated summaries for several models and the reference on the CNN/DM dataset.}
        \label{fig:density_word_en}
    \end{subfigure}
    \caption{Information about the length of the generated summaries for the CNN/DM dataset.}
    \label{fig: info_gen_en}
\end{figure*}

On the CNN/DailyMail dataset, STRASS tends to extract less sentences but longer ones comparing to the \(Oracle sent\) (figure \ref{table:en_result_len}, \ref{fig:density_sent_en}, \ref{fig:density_word_en}). On the figure \ref{fig:pos_sum20_en_model} we can see that the three models tend to extract different sentences. \(Oracle sent\) which is the best performing model tends to extract the 4 first sentences, \(Oracle\) extracts more often the fourth sentences than the first three and still have better results than the \(Lead3\), which means that the fourth sentences could have some interest. With STRASS the first three sentences have a different tendency than the rest of the text, showing that the first three sentences may have a different structure than the rest. Then, the farther a sentence is in the text, the lower the probability to take it.

%\section{Supplemental Material}
%\label{sec:supplemental}
%Submissions may include non-readable supplementary material used in the work and described in the paper. Any accompanying software and/or data should include licenses and documentation of research review as appropriate. Supplementary material may report preprocessing decisions, model parameters, and other details necessary for the replication of the experiments reported in the paper. Seemingly small preprocessing decisions can sometimes make a large difference in performance, so it is crucial to record such decisions to precisely characterize state-of-the-art methods. 

%Nonetheless, supplementary material should be supplementary (rather
%than central) to the paper. \textbf{Submissions that misuse the supplementary 
%material may be rejected without review.}
%Supplementary material may include explanations or details
%of proofs or derivations that do not fit into the paper, lists of
%features or feature templates, sample inputs and outputs for a system,
%pseudo-code or source code, and data. (Source code and data should
%be separate uploads, rather than part of the paper).

%The paper should not rely on the supplementary material: while the paper
%may refer to and cite the supplementary material and the supplementary material will be available %to the
%reviewers, they will not be asked to review the
%supplementary material.

\end{document}